\documentclass[conference]{IEEEtran}
\IEEEoverridecommandlockouts

\usepackage{amsmath,amssymb,amsfonts}
\usepackage{algorithmic}
\usepackage{graphicx}
\usepackage{textcomp}
\usepackage{xcolor}
\usepackage{graphicx}
\usepackage{fancyhdr}
\usepackage{subfigure}
\usepackage{nidanfloat}
\graphicspath{{./images/}} 
\usepackage[caption=false]{subfig}
\def\BibTeX{{\rm B\kern-.05em{\sc i\kern-.025em b}\kern-.08em
    T\kern-.1667em\lower.7ex\hbox{E}\kern-.125emX}}
\begin{document}

\title{Thermal to Visible Face Recognition Using Deep Autoencoders}
\renewcommand{\refname}{References}
\author{\IEEEauthorblockN{Alperen Kantarc{\i}}
\IEEEauthorblockA{\textit{Department of Computer Engineering} \\
\textit{Istanbul Technical University}\\
Istanbul, Turkey\\
kantarcia@itu.edu.tr}
\and
\IEEEauthorblockN{Haz{\i}m Kemal Ekenel}
\IEEEauthorblockA{\textit{Department of Computer Engineering} \\
\textit{Istanbul Technical University}\\
Istanbul, Turkey\\
ekenel@itu.edu.tr}
}

\maketitle

\begin{abstract}
Visible face recognition systems achieve nearly perfect recognition accuracies using deep learning. 
However, in lack of light, these systems perform poorly. A way to deal with this problem is thermal to visible cross-domain face matching. This is a desired technology because of its usefulness in night time surveillance. Nevertheless, due to differences between two domains, it is a very challenging face recognition problem. In this paper, we present a deep autoencoder based system to learn the mapping between visible and thermal face images. 
Also, we assess the impact of alignment in thermal to visible face recognition. For this purpose, we manually annotate the facial landmarks on the Carl and EURECOM datasets. The proposed approach is extensively tested on three publicly available datasets: Carl, UND-X1, and EURECOM. Experimental results show that the proposed approach improves the state-of-the-art significantly. We observe that alignment increases the performance by around 2\%. Annotated facial landmark positions in this study can be downloaded from the following link: github.com/Alpkant/Thermal-to-Visible-Face-Recognition-Using-Deep-Autoencoders .
\end{abstract}


\section{Introduction}
Face recognition has been a popular research area for both military and commercial purposes. Most of the face recognition technologies use visible spectrum images. In the past few years, many studies presented very high recognition accuracies. However, in the night time, thermal images are better options because thermal cameras capture heat waves of the objects without any illumination source. Night time surveillance is very crucial for the police forces and military operations in order to recognize the suspects. Most of the visible face recognition methods give poor performance under the illumination changes and low lighting. Therefore, face images that are taken at the night time lead to poor face recognition accuracies. Cross-domain face recognition is used between thermal and visible images to overcome this limitation. In thermal to visible face recognition, the task is to match the thermal probe image of a person with the visible image of the same person within the visible gallery. It is a challenging problem to match faces between thermal and visible domains due to the difference in appearance characteristics of these domains. Moreover, most thermal cameras capture low resolution images, whereas visible images are high resolution, and this increases the domain gap. %Because of the modality difference, it is one of the most challenging face matching problem. This challenging nature of the problem and the limited number of research in this area creates a significant research gap. 

 %The visible gallery contains different images of the different identities, similar to the police mugshot database.

In this paper, we aim at learning a mapping between thermal and visible domains using a deep autoencoder model. We use three thermal-visible datasets to evaluate the performance of the proposed system. In addition, we manually annotated six facial landmarks on the Carl and EURECOM datasets in order to align the faces and assess alignment's impact on the performance. Our experiments show around 14\% and 3.5\%  absolute increase in rank-1 accuracies on Carl \cite{carl2010,carl2013} and UND-X1 \cite{notredame,notredame2} datasets, respectively. We also present thermal to visible face recognition results on the recently collected EURECOM dataset \cite{eurecom}.  

The contributions of this paper can be summarized as follows: First, we have shown that deep convolutional autoencoders can learn non-linear mapping between thermal and visible images for cross domain face recognition task. Second, we analyze the impact of alignment for cross domain face recognition. Moreover, these produced annotations can also be used to develop facial landmark detection methods for thermal face images. Finally, we have improved the state-of-the-art face recognition accuracies on publicly available thermal-visible face datasets. %Third, we manually labeled facial landmarks of the faces which is needed for the face alignment. Also, we will share these landmark information with the dataset owners for future research. 

The remainder of this paper is organized as follows. In Section 2, we give a brief overview of the related work. Then, in Section 3, we explain the utilized model and the approach. In Section 4, details of the datasets, benchmarks, and experimental results are presented and discussed. The last section concludes the paper and gives directions for the future work.

\section{Related Work}

One of the initial successful works before deep learning approaches for thermal to visible matching is presented in \cite{partialleastsquares}. %research is one of the most successful one until the neural network approaches developed. 
In this study, the authors utilized partial least squares method. They investigated the effects of preprocessing and domain gap very profoundly. Mean filtering, geometric normalization, and Difference of Gaussians (DoG) filtering methods were employed and compared. We benefited from the proposed preprocessing methods in this paper.

%Unlike the NIR and shortwave infrared bands (SWIR), mid wave (MWIR) and long wave infrared (LWIR) bands are completely passive, which means that they do not require active illumination and methods used in these bands similar to each other \cite{recentadvances}. The most successful research for this band are \cite{deepperceptual} and \cite{autoassociative} that use direct and indirect regression respectively.

One of the first and most successful approaches to the thermal to visible face recognition using neural networks is Deep Perceptual Mapping (DPM) \cite{deepperceptual}. In the proposed system, the authors aim at learning a mapping between the thermal and visible spectrum face images. Their simple feed forward neural network uses densely computed hand-crafted feature vectors. %DPM also performed the state of the art recognition accuracies on the UND-X1 and Carl datasets which we increased these accuracies in this paper. 
Deep Perceptual Mapping network has been trained with the overlapping patch crops of the faces, which correspond to the same area of the face both in thermal and visible domain. Different from this study, in this paper, we use whole face images instead of small patches of the face, the feature vectors are generated using convolutional neural networks (CNN), and the mapping is learned using a deep convolutional autoencoder. %Preprocessing methods are similar but details have given in the following sections.

After the success of generative adversial networks (GANs) in image synthesis, researchers employed GANs to the thermal to visible face recognition task. TV-GAN \cite{TVGAN} is one of the most successful GAN based approaches to the problem. Aim of the TV-GAN is to transform thermal faces to visible light faces while preserving the identity information. Authors trained the discriminator to perform both binary classification of real or fake and perform closed-set face recognition task. 

SG-GAN \cite{SGGAN} synthesizes visible images from thermal images by regularizing the GAN training with semantic priors which are extracted by a face parsing network. Main novelty of this work is semantic loss function employed to regularize the training process to reduce the per-pixel loss value calculated between the synthesized visible image and thermal image.

\section{Proposed Approach}
To learn the mapping between the visible and thermal face images, we utilized a deep autoencoder architecture. Specifically, we benefited from the U-Net architecure \cite{UNetCN} and modified it for the purposes of this study.

%Here, we first describe U-Net \cite{UNetCN} then we share the details of the model that we used. The used network is based on U-Net but it is slightly modified.

Autoencoders are special types of neural networks that learn data embeddings in an unsupervised manner. Autoencoders are useful to encode the data to the latent code and retrieve the original data using this latent code, since their inputs are the same with their outputs. They have been used for different purposes, such as denoising, dimensionality reduction, and image reconstruction. %The task of the autoencoder is regression where the network is asked to predict its input, but use cases increase when input and output are not the same. 
One of the most well-known autoencoder architectures is the U-Net~\cite{UNetCN}, which is proposed for the medical segmetation task. 

\begin{figure*}[t]
	\centering
	\includegraphics[width=\textwidth]{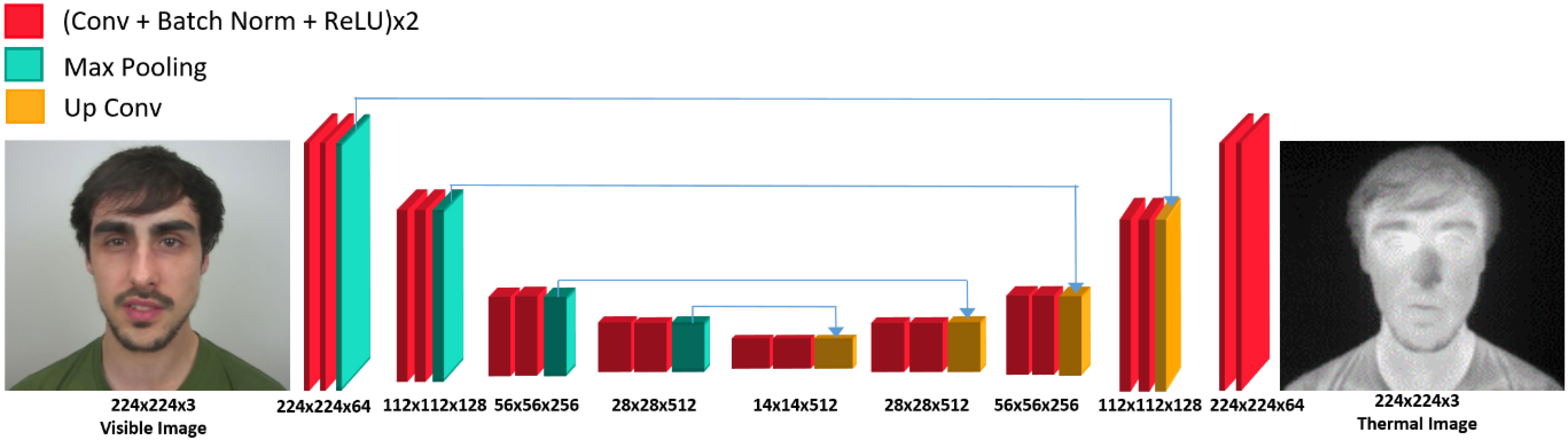}
	\caption{\label{fig:arch-illustration}Illustration of the network architecture }
\end{figure*}

Our network is based on the U-Net architecture~\cite{UNetCN}. U-Net uses 1024 channels and $30\times30$ bottleneck layer. Since the amount of training samples is limited in thermal to visible face recognition, in order to decrease the number of trainable parameters, we use less channels. Our model has 512 channels and $14\times14$ pixels in the bottleneck layer. In the proposed approach, input images are visible face images and outputs  are  the face images from the LWIR band to learn a mapping between visible face images and thermal face images. The network takes $224\times224$ pixel resolution color images and the output thermal image has the same size as the input. Convolution filter size is $3\times3$ and window size of the max pooling is two, which is the same size as in U-Net~\cite{UNetCN}. Activation function after each batch normalization layer is Rectified Linear Unit (ReLU). For the decoder, we use two different upsampling methods. First one is bilinear upsampling, which is a standard interpolation technique. As the number of trainable parameters decrease by using the bilinear upsampling, required GPU memory reduces and training time decreases. However, information for the decoding part is lost and it degrades the performance. Therefore, besides bilinear upsampling, we also employ up convolution with $2\times2$ filter size as proposed in the U-Net \cite{UNetCN}. The number of total trainable parameters of the up convolution version is 14,789,059. The loss function for this network is mean square loss, since the network tries to make its output as similar as possible to the ground truth thermal image. Fig. \ref{fig:arch-illustration} shows the proposed architecture.

\section{Datasets and Experiments}
This section presents the datasets that have been used to assess the performance of the proposed approach, the experiments carried out, and the obtained results.
\subsection{Datasets}
Three thermal-visible datasets have been used in this paper. All of them are paired LWIR-visible face datasets. For the Carl and EURECOM datasets, we manually annotated six facial landmark points: two corners for each eye and mouth corners. %Manual alignment is needed for the thermal face images, because the accuracy of the automated facial landmark detection systems are low unlike the accuracy of visible light domain face landmark detection systems. 

\textbf{Carl Dataset:} Carl dataset \cite{carl2010,carl2013} contains 41 different subjects with NIR, LWIR, and visible paired images. We only use LWIR and visible images. In total, there are 4920 images, half of them are visible images and the other half are LWIR thermal images. We use the same benchmark settings proposed in \cite{deepperceptual} in terms of train and test subject ratio. That is, we used images of 20 randomly chosen subjects as training data and the remaining 21 subjects as testing data. Training and testing subjects are completely different. %and testing subjects that have never seen before by the network. 
All available thermal images of 21 subjects have been used as thermal probes which sums to 1260. There are three different visible gallery settings that have been used: one image per subject, two images per subject, and all available images of the subjects. So visible gallery contains 21, 42, and 1260 images, respectively. %These three settings show the performance of the model under challenging and common scenarios.     

\textbf{UND-X1 Collection:} University of Notre Dame X1 Collection dataset \cite{notredame,notredame2} contains 82 different subjects with  LWIR and visible paired images. There are different expressions and images collected in different sessions. We follow the same benchmarking protocol presented in \cite{partialleastsquares} and \cite{deepperceptual}. Training set contains 41 subjects and testing set contains remaining 41 subjects. However, for each subject there are different number of images in this dataset, unlike the other datasets. Visible gallery settings are the same with the ones that we use on the Carl dataset. For each setting the number of total visible images are 41, 82, and 1146. 

\textbf{EURECOM Dataset:} EURECOM dataset has been recently presented in \cite{eurecom}. It contains seven different facial variations, five different illumination conditions, five different types of occlusions, and four different head poses that belong to 50 subjects. In total, there are 2100 images in the dataset. We manually marked six facial landmarks also on this dataset. However, it is impossible to annotate some facial landmarks in different variations, therefore we excluded some of them. More specifically, we excluded the following variations: all lights off, eyeglasses, sunglasses, mouth occluded by hand, eye occluded by hand, and poses with 30$^{\circ}$ left and right. In the end, for each subject, we have twelve images from each domain that sums up to 1200 images. Since we excluded some images and number of available images is relatively smaller than the other datasets, we chose 30 subjects to the training set and used remaining 20 subjects for testing. For testing, all available thermal images of the subjects are used as thermal probes, that sums up to 240 thermal images. For visible gallery image per subject settings are the same with the Carl and UND-X1 datasets. For the three settings 20, 40, and 240 images are used in the gallery, respectively.
\begin{figure*}[t]
	\centering
	\includegraphics[width=\linewidth]{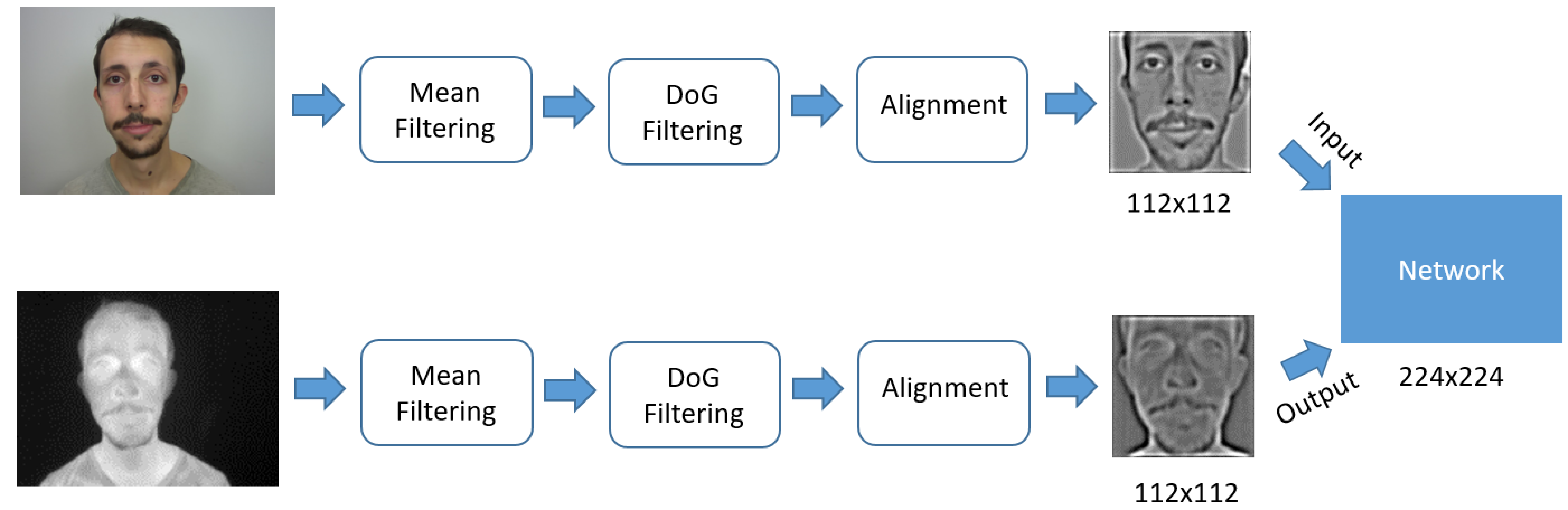}
	\caption{\label{fig:preprocessing}Preprocessing steps }
\end{figure*}
\subsection{Preprocessing}
Preprocessing of the images is an important task in thermal to visible face recognition. Due to the significant domain gap between the thermal and visible images, preprocessing methods would help to close the gap between these two domains and make it easier to learn the mapping between them. All the images pass through the same preprocessing steps shown in Fig. \ref{fig:preprocessing}. First $3\times3$ mean filter is applied in order to remove the dead pixels in the thermal images, then DoG filter is applied to enhance the edges and then both visible and thermal images are aligned according to the manually annotated facial landmarks. Because of the resolution difference between thermal images and visible images, both thermal and visible images are downsampled to $112\times112$ pixels which is closer to the resolution of the thermal image. Downsampling the high resolution image to lower resolution to match with the thermal image resolution helps to decrease the domain gap between two images, since resolution difference decreases between two images. Then both images are upsampled to $224\times224$, which is input and output resolution of the CNN. We have also tested the network without applying DoG filter and the results are presented in the following section.

\subsection{Experiments}
As a first step, we perform preprocessing on the images. Then for each dataset, these processed thermal and visible pairs are given as input and output to the deep autoencoder architecture, respectively. Initial learning rate is 0.01 with Adam optimizer. Learning rate is halved when validation error does not change multiple epochs. Batch size is 32 for all the experiments. Training time changes according to the dataset size. Average training time takes five hours with Tesla K80 GPU. 

To perform face recognition, each gallery image has been given to the network as input and the network generated corresponding thermal image. Then each thermal probe image has been compared with these thermal images generated by the network.

We conduct experiments with three different setups. In the first two experiments we investigate the effect of the upsampling method for the decoder. For these experiments, DoG filter were not applied to images as preprocessing step. For the third experiment, we test the effect of the DoG filter on the upconvolutional autoencoder model. %The proposed model is the model with up convolution with DoG filtered images. These three experiments have been made on the three datasets. 
In order to make a fair comparison with the \cite{deepperceptual}, we also use face images that are not aligned on the Carl dataset. Results are presented as averaged rank-1 recognition accuracies over ten independent experiments. That is, for each dataset, each model is trained and tested ten different times and the obtained results are averaged.

\subsection{Results}

Table 1, 2, and 3 present the averaged rank-1 recognition scores on the Carl, UND-X1, and EURECOM datasets, respectively.
%Performance of the network is calculated as described in the Section 1. 
In the tables, we also include the accuracies of the state-of-the-art methods for comparison purposes. For the Carl and UND-X1 datasets, state-of-the-art models are Deep Perceptual Mapping \cite{deepperceptual} and Hu et al. \cite{partialleastsquares}. %which is the former state of the art method. 
On EURECOM dataset, only \cite{crossspectrumeurecom} presented a study for the thermal to visible recognition. The authors propose an image synthesis method by using cascaded refinement network. However, their benchmark setup is different than ours and earlier thermal to visible face recognition studies. %They evaluate the performance by comparing generated visible-like faces against the visible gallery. 
%Also, they compare the accuracies with respect to different face variations available in the dataset. %As we have described before, we excluded some face variations to make face landmark annotations for face alignment. 
%They use only five subjects for testing. They obtained 82\% accuracy under neutral face variation and on average 57.61\% under all the available face variations. %We could not implement the \cite{deepperceptual} method on the eurecom dataset to compare the results with ours due to project requirements of the \cite{deepperceptual}. Therefore, we only share the our accuracy scores on the eurecom dataset. In future work we would like to see the comparison of our methods with the others in the same benchmark on the eurecom dataset.

\begin{table*}[tb]
\caption{\label{tbl:carl}Rank-1 recognition accuracies on the Carl dataset}
\centering
\begin{tabular}{lccc}
\multicolumn{1}{c}{Models} & \multicolumn{3}{c}{\#Visible Image Gallery/Subject} \\ \hline
 & 1/subject & 2/subject & all/subject \\ \cline{2-4} 
Deep Perceptual Mapping \cite{deepperceptual} & 56.33\% & 60.08\% & 71\% \\
Partial Least Squares \cite{partialleastsquares} & 31.75\% & 34.66\% & 51.58\% \\
Model with Bilinear Upsample  & 40.2\% & 45.8\% & 75.5\% \\
Model with Bilinear Upsample (Aligned) & 42\% & 48.75\% & 77.25\% \\
Model with Up Convolution  & 41\% & 49.75\% & 80.2\% \\
Model with Up Convolution (Aligned) & 43.75\% & 52.5\% & 82.5\% \\
\textbf{Model with Up Convolution + DoG Filter } & \textbf{46.8\%} & \textbf{58.5\%} & \textbf{82.5\%}\\
\textbf{Model with Up Convolution + DoG Filter (Aligned)} & \textbf{48\%} & \textbf{60.25\%} & \textbf{85\%}
\end{tabular}
\end{table*}

The highest accuracies are reached when preprocessing, up convolution, and alignment are used. As can be seen in Table \ref{tbl:carl}, the proposed method is superior to the DPM, when two visible images per subject and all available visible images per subject are used in the gallery. Table \ref{tbl:undx1} shows that our proposed method outperforms DPM on the UND-X1 dataset in all the gallery settings. We improved the state-of-the-art by an absolute increase of 14\% in accuracy on the Carl dataset and by 3.5\% on the UND-X1 dataset when all the samples of the subjects are used in the gallery. From Table \ref{tbl:carl}, it can also be observed that alignment improves the performance by around 2\%. Up convolution is found to be more useful compared to upsampling leading to increased accuracies on all there datasets. Similarly, applying preprocessing has enhanced the performance further on all three datasets, which indicates its importance. Please note that the results presented in Table \ref{tbl:eurecom} are not directly comparable with the ones presented in \cite{crossspectrumeurecom}. In \cite{crossspectrumeurecom} authors use 45 subjects for training and 5 subjects for testing, whereas we use 30 subjects for training and 20 subjects for testing. Considering that we use less amount of training samples and have more classes to discriminate, our setup poses a more challenging problem.   % We also used the the provided test images which are not aligned with our manual landmarks on Carl dataset. These images are provided by the authors of the dataset and \cite{deepperceptual} use these images. In order to make better comparison with our work, we used both non aligned as \cite{deepperceptual} did and aligned according to our manual landmarks. These results also indicate that our best model perform better than \cite{deepperceptual} when all visible images used as gallery image.

\begin{table}[b]
\caption{\label{tbl:undx1}Rank-1 recognition accuracies on the UND-X1 dataset}
\centering
\resizebox{0.5\textwidth}{!}{
\begin{tabular}{lccc}
\multicolumn{1}{c}{Models} & \multicolumn{3}{c}{\#Visible Image Gallery/Subject} \\ \hline
 & 1/subject & 2/subject & all/subject \\ \cline{2-4} 
Deep Perceptual Mapping \cite{deepperceptual} & 55.36\% & 60.83\% & 83.73\% \\
Partial Least Squares \cite{partialleastsquares} & 44.75\% & 50.89\% & 69.86\% \\
Model with Bilinear Upsample & 42\% & 50.25\% & 75.4\% \\
Model with Up Convolution & 49.25\% & 57.5\% & 82\% \\
\textbf{Model with Up Convolution + DoG Filter} & \textbf{58.75\%} & \textbf{65.25\%} & \textbf{87.2\%}
\end{tabular}%
}
\end{table}

% In the gallery setting with all available visible images of the subject, we improved the state of the art by 4\%. 

\begin{table}[b]
\caption{\label{tbl:eurecom}Rank-1 recognition accuracies on the EURECOM dataset}
\centering
\resizebox{0.5\textwidth}{!}{
\begin{tabular}{lccc}
\multicolumn{1}{c}{Models} & \multicolumn{3}{c}{\#Visible Image Gallery/Subject} \\ \hline
 & 1/subject & 2/subject & all/subject \\ \cline{2-4} 
Model with Bilinear Upsample & 50.41\% & 58.5\% & 80\% \\
Model with Up Convolution & 53.75\% & 61.25\% & 81.25\% \\
\textbf{Model with Up Convolution + DoG Filter} & \textbf{57.91\%} & \textbf{70\%} & \textbf{88.33\%}
\end{tabular}%
}
\end{table}

%The only thermal to visible matching study conducted on the eurecom dataset is \cite{crossspectrumeurecom} which showed 52\% Rank-1 accuracy on average and 82\% on neutral face variation on the five subjects. Although our proposed method gets higher recognition accuracy than their best results, we cannot directly compare two methods because benchmarks of \cite{crossspectrumeurecom} and ours are different.

%Results indicate that using up convolution increase the performance significantly. Also, it points out the importance of the DoG filter and effect of the closed modality gap by the preprocessing methods. It is clear that Rank-1 accuracies are higher than the state of the art methods in Carl and UND-X1 datasets. 

\begin{figure}[t]
    \centering
    \subfigure[Correctly matched subjects]{
    \includegraphics[height=3.5cm]{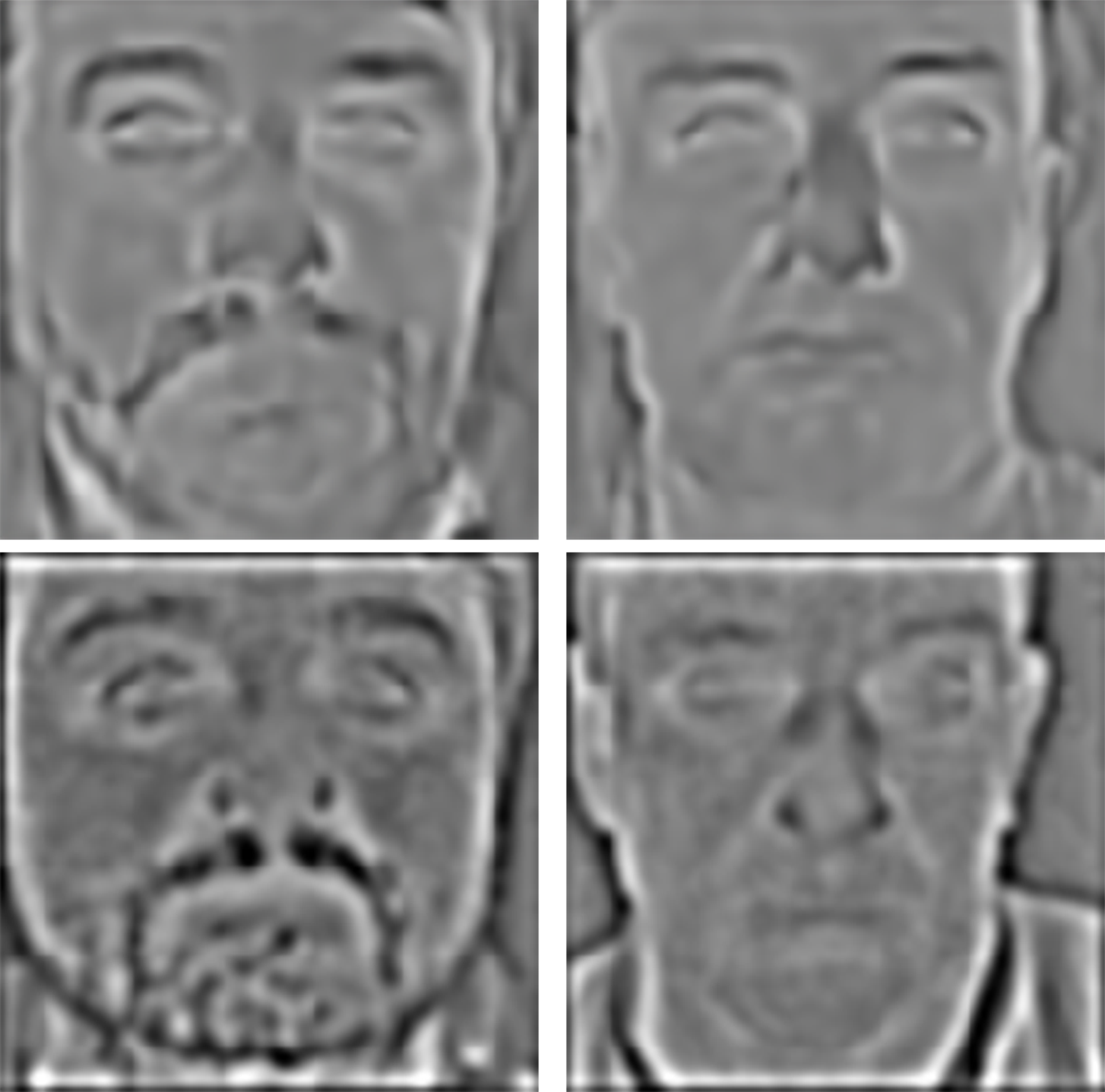}}
\qquad
\subfigure[Mismatched subjects]{
\includegraphics[height=3.5cm]{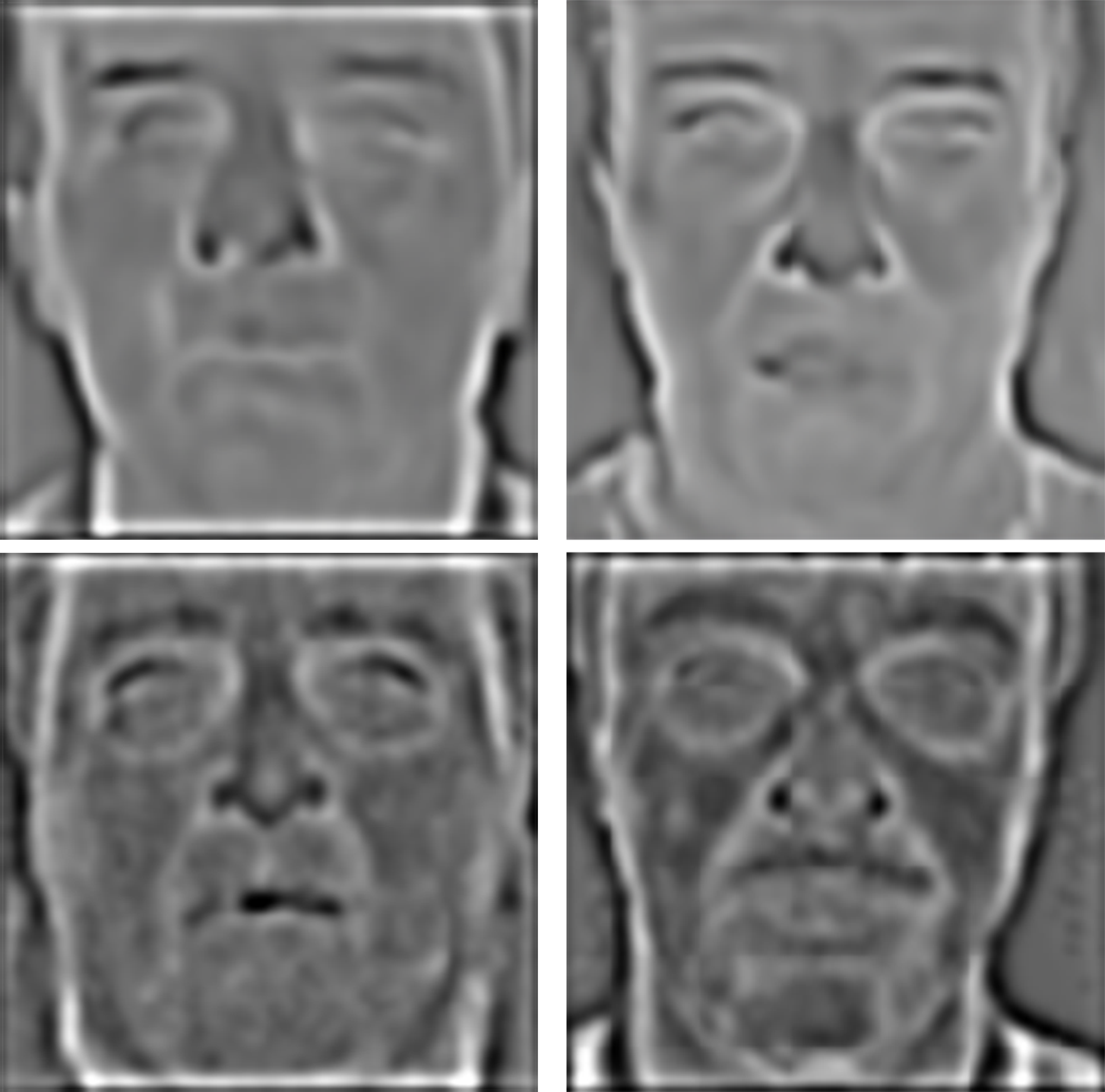}}
	\caption{\label{fig:modeloutput} Top row shows the outputs of the model and bottom row shows the ground truth thermal images of the subjects. Correctly predicted subject (a) shows visually satisfying outputs and (b) shows the mismatched subject images. }
\end{figure}

Fig. \ref{fig:modeloutput} shows correctly matched and mismatched samples. Top row contains the outputs of the autoencoder and bottom row contains the ground truth thermal images of the subjects. As can be seen, for the correct matches, the autoencoder provides visually satisfying results. However, as in the case of mismatches, if the autoencoder fails to generate a good mapping, then consequently the classification system also fails.

%Correctly predicted subject (a) shows the visually satisfying results and (b) shows the mismatched subject images. 

%in order to visualize the model's performance on the images.

\section{Conclusion}
In this paper, we show that mapping between thermal and visible images can be learned successfully by using a deep convolutional autoencoder. In addition, we show that applying preprocessing and alignment improves the performance further. We improve the state-of-the-art face recognition accuracies on two different and widely used datasets and also present results on a recently collected dataset. As a future work, we will explore different deep convolutional autoencoder architectures. We also plan to address several appearance variations, such as occlusion and head pose, in thermal face images. %Moreover, a deeper autoencoder can be trained with more training data which could increase the performance of the model.

\bibliographystyle{IEEEtran}
\bibliography{references}

\begin{thebibliography}{10}
\providecommand{\url}[1]{#1}
\csname url@samestyle\endcsname
\providecommand{\newblock}{\relax}
\providecommand{\bibinfo}[2]{#2}
\providecommand{\BIBentrySTDinterwordspacing}{\spaceskip=0pt\relax}
\providecommand{\BIBentryALTinterwordstretchfactor}{4}
\providecommand{\BIBentryALTinterwordspacing}{\spaceskip=\fontdimen2\font plus
\BIBentryALTinterwordstretchfactor\fontdimen3\font minus
  \fontdimen4\font\relax}
\providecommand{\BIBforeignlanguage}[2]{{%
\expandafter\ifx\csname l@#1\endcsname\relax
\typeout{** WARNING: IEEEtran.bst: No hyphenation pattern has been}%
\typeout{** loaded for the language `#1'. Using the pattern for}%
\typeout{** the default language instead.}%
\else
\language=\csname l@#1\endcsname
\fi
#2}}
\providecommand{\BIBdecl}{\relax}
\BIBdecl

\bibitem{carl2010}
\BIBentryALTinterwordspacing
V.~Espinosa-Dur{\'o}, M.~Faundez-Zanuy, J.~Mekyska, and E.~Monte-Moreno, ``A
  criterion for analysis of different sensor combinations with an application
  to face biometrics,'' \emph{Cognitive Computation}, vol.~2, no.~3, pp.
  135--141, Sep 2010. [Online]. Available:
  \url{https://doi.org/10.1007/s12559-010-9060-5}
\BIBentrySTDinterwordspacing

\bibitem{carl2013}
V.~Espinosa-Dur{\'o}, M.~Faundez-Zanuy, and J.~Mekyska, ``A new face database
  simultaneously acquired in visible, near-infrared and thermal spectrums,''
  \emph{Cognitive Computation}, vol.~5, pp. 119--135, 03 2013.

\bibitem{notredame}
X.~Chen, P.~J.~Flynn, and K.~Bowyer, ``Visible-light and infrared face
  recognition,'' in \emph{Workshop on Multimodal User Authentication}, 2003,
  p.~48.

\bibitem{notredame2}
P.~J.~Flynn, K.~Bowyer, and P.~J. Phillips, ``Assessment of time dependency in
  face recognition: An initial study,'' in \emph{International Conference on
  Audio-and Video-Based Biometric Person Authentication}.\hskip 1em plus 0.5em
  minus 0.4em\relax Springer, 2003, pp. 44--51.

\bibitem{eurecom}
K.~{M}allat and J.~{D}ugelay, ``A benchmark database of visible and thermal
  paired face images across multiple variations,'' in \emph{International
  Conference of the Biometrics Special Interest Group, {BIOSIG} 2018,
  Darmstadt, Germany, September}, 2018, pp. 199 -- 206.

\bibitem{partialleastsquares}
S.~Hu, J.~Choi, A.~L. Chan, and W.~R. Schwartz, ``Thermal-to-visible face
  recognition using partial least squares,'' \emph{JOSA A}, vol.~32, no.~3, pp.
  431--442, 2015.

\bibitem{deepperceptual}
\BIBentryALTinterwordspacing
M.~S. Sarfraz and R.~Stiefelhagen, ``Deep perceptual mapping for cross-modal
  face recognition,'' \emph{Int. J. Comput. Vision}, vol. 122, no.~3, pp.
  426--438, May 2017. [Online]. Available:
  \url{https://doi.org/10.1007/s11263-016-0933-2}
\BIBentrySTDinterwordspacing

\bibitem{TVGAN}
T.~{Zhang}, A.~{Wiliem}, S.~{Yang}, and B.~{Lovell}, ``Tv-gan: Generative
  adversarial network based thermal to visible face recognition,'' in
  \emph{2018 International Conference on Biometrics (ICB)}, Feb 2018, pp.
  174--181.

\bibitem{SGGAN}
C.~{Chen} and A.~{Ross}, ``Matching thermal to visible face images using a
  semantic-guided generative adversarial network,'' in \emph{2019 14th IEEE
  International Conference on Automatic Face Gesture Recognition (FG 2019)},
  May 2019, pp. 1--8.

\bibitem{UNetCN}
O.~Ronneberger, P.~Fischer, and T.~Brox, ``U-net: Convolutional networks for
  biomedical image segmentation,'' in \emph{MICCAI}, 2015.

\bibitem{crossspectrumeurecom}
K.~{Mallat}, N.~{Damer}, F.~{Boutros}, A.~{Kuijper}, and J.~{Dugelay},
  ``Cross-spectrum thermal to visible face recognition based on cascaded image
  synthesis,'' in \emph{2019 International Conference on Biometrics (ICB)},
  June 2019.

\end{thebibliography}

\end{document}